# A Locally Statistical Active Contour Model for Image Segmentation with Intensity Inhomogeneity

Kaihua Zhang, Lei Zhang[*], Kin-Man Lam, and David Zhang

***Abstract*** —A novel locally statistical active contour model (ACM) for image segmentation in the presence of intensity inhomogeneity is presented in this paper. The inhomogeneous objects are modeled as Gaussian distributions of different means and variances, and a moving window is used to map the original image into another domain, where the intensity distributions of inhomogeneous objects are still Gaussian but are better separated. The means of the Gaussian distributions in the transformed domain can be adaptively estimated by multiplying a bias field with the original signal within the window. A statistical energy functional is then defined for each local region, which combines the bias field, the level set function, and the constant approximating the true signal of the corresponding object. Experiments on both synthetic and real images demonstrate the superiority of our proposed algorithm to state-of-the-art and representative methods.

# I. INTRODUCTION

Image segmentation is an important procedure in many computer vision and pattern recognition applications. Many promising methods have been proposed for image segmentation, such as the region merging based methods [1-4], the graph based methods [5-8], and the active contour model (ACM) based methods [9-10][13-25][35-36][41-42], etc. ACM for segmentation aims to drive the curves to reach the boundaries of the interested objects. The driven forces are mainly from the image data, including edge-based [16][42] or region-based forces [13-15][17-21][23-25][41]. The edge-based ACM often utilizes the local image gradient information to build some stopping functions in order to drive the contour to stop at the object boundary, while the region-based ACM aims to drive the curve to evolve through some region-based descriptors [21-25]. The edge-based ACM methods are applicable to images with intensity inhomogeneity but, in general, they are sensitive to the initialization of the level set function. Moreover, they easily suffer from serious boundary leakage for images with weak boundaries [19]. Many region-based ACMs are based on the global region information; they assume that the image intensity is homogeneous [13][22], and thus are not suitable for segmenting images with intensity inhomogeneity.

Intensity inhomogeneity caused by the imperfection of imaging devices or by illumination variations often occurs in real-world images, and it can lead to serious misclassifications by intensity-based segmentation algorithms that assume a uniform intensity [13][22]. Statistically, misclassifications are caused by the prolonged tails of the intensity distribution of each object so that it is difficult to extract the desired objects accurately based on their respective intensity distributions (refer to Fig. 1 please). The well-known Mumford-Shah (MS) model [12], which assumes that an image is piecewise smooth, is suitable for modeling images with intensity inhomogeneity. The MS model uses a set of contours $C$ to separate different regions. However, it is difficult to minimize the energy functional of the MS model because the set $C$ of low dimension is unknown and the problem is non-convex [14]. Some simplified versions of the MS model have been proposed, such as the seminal Chan-Vese (CV) model [13] and the piecewise-smoothing (PS) model [14-15]. All these methods represent the contour $C$ using the zero level of a function called the level set function, and then segmentation proceeds by evolving a level set evolution equation, obtained by minimizing some energy functional. However, the CV model is not applicable to images with intensity inhomogeneity because it models images by means of the piecewise constant. The PS model can yield a desirable segmentation for images with intensity inhomogeneity. However, the iteration of two partial differential equations (PDE) is needed in the PS model to approximate the original image, and this limits the practical application of PS model because it is very time-consuming.

Recently, some local region-based ACMs have been proposed for images with intensity inhomogeneity, such as the local region descriptors (LRD) method [20], the local region based (LRB) method [21], the local binary fitting (LBF) model [17-18], the local intensity clustering (LIC) method [19], and the local region model (LRM) [41], etc. However, there exist some drawbacks with these local region based models. The LRD model needs to tune a balloon force parameter for images with strongly overlapping intensity distribution, but how to define the degree of the overlap is not mentioned in [20]. Therefore, it is difficult to select a proper balloon force. The LRB method has two drawbacks. First, the Dirac functional used there is restricted to a neighborhood around the zero level set, which makes the level set evolution act locally. Therefore, the evolution can be easily trapped in local minima [13]. Second, the region descriptor in LRB is only based on the region mean information without considering the region variance, and this may lead to inaccurate segmentation. The second drawback of LRB also holds for the LBF model, because they use a similar energy functional. The LIC method can be considered as a locally weighted $K$-means clustering method [19]. It does not consider the clustering variance, which may cause inaccurate segmentation; similar drawback exists for the $K$-means clustering based method in [26]. The LRM method relates the local region statistics, i.e., local region means and variances, in interpreting the MS model. However, the local region means and variances are only defined empirically, but not derived from minimizing the MS energy.

Intensity inhomogeneity is usually ascribed to a smooth and spatially varying field multiplying the constant true signal of the same object in the measured image [11][19]. This spatially varying and smooth field is called a bias field. This paper presents a novel statistical ACM for simultaneous segmentation and bias correction. By exploiting the image's local redundant information, we define a mapping from the original image domain to another domain such that the intensity probability model is more robust to noise and the overlapping of intensity is suppressed to some extent. We then devise a statistical energy functional for the distribution of each local region in the transformed domain, which combines the bias field, the level set function, and the constant approximating the signal of the corresponding object. Analysis of the proposed approach shows that it is a *soft classification* model, which means that each pixel can be assigned to more than one class. In contrast, the *hard classification* used in previous methods [13][45] assigns each pixel to only one class. Therefore, the proposed approach can achieve a better segmentation result. In addition, the proposed method can be applied to simultaneous tissue segmentation and bias correction for magnetic resonance (MR) images. Our preliminary work on this has been briefly presented in [23].

The rest of the paper is organized as follows. Section II introduces the background of our research, including the seminal MS model, the CV model, and the PS model. Section III describes the statistical model for intensity inhomogeneity and our proposed model in detail. Section IV presents the proposed algorithm. Section V shows extensive experimental results. In Section VI, we discuss the relationships between our method and the CV model, the LBF model and the LIC model. Section VII concludes the paper.

## II. BACKGROUND

Let $\Omega \subset R^N$, where $N = 2$ or 3, be the image domain and $I(\mathrm{x}): \Omega \rightarrow R$ be an input image. Mumford and Shah [12] approximated an image with a piecewise-smooth function $u(\mathrm{x}): \Omega \rightarrow R$, such that $u$ varies smoothly within each of the sub-regions, and abruptly across the boundaries of the sub-regions. Let $C(p): R \rightarrow \Omega$ approximate the edges of the sub-regions. The energy functional is given as follows [12]:

$$E^{MS}(u,C) = \int_{\Omega}(I-u)^2 d\mathrm{x} + \mu \int_{\Omega \backslash C}|\nabla u|^2 d\mathrm{x} + \nu |C| \qquad (1)$$

where $\mu, \nu > 0$ are two fixed parameters, and $|C|$ represents the length of the contour. The image segmentation can be performed by minimizing Eq.(1) with respect to $u$ and $C$. However, it is difficult to minimize the above functional in practice, due to the unknown set $C$ of lower dimension and the non-convexity of the functional [14]. Many methods have been proposed to simplify or modify the functional [13-15], which will be reviewed in the following. Here we mainly discuss two-phase cases, where the region $\Omega$ is separated by a contour $C$ and $\Omega = \Omega_{inside(C)} \cup \Omega_{outside(C)}$, $\Omega_{inside(C)} \cap \Omega_{outside(C)} = \varnothing$. The results can be readily extended to multi-phase cases by utilizing two different level set functions to represent different regions (refer to [14] for details).

Chan and Vese [13] proposed an active contour model which can be seen as a particular case of Eq.(1), and they represented $u(\mathrm{x}): \Omega \rightarrow R$ in Eq.(1) as a piecewise-constant function. Chan and Vese proposed to minimize the following energy functional (i.e., the CV model)

$$E^{CV}(c_1, c_2, C) = \int_{inside(C)}(I-c_1)^2 d\mathrm{x} + \int_{outside(C)}(I-c_2)^2 d\mathrm{x} + \nu |C| \qquad (2)$$

where $c_1$ and $c_2$ are two constant functions which approximate the average intensities inside and outside the contour $C$, respectively. This energy functional can be represented by a level set formulation, and by evolving the level set, we can obtain its minimum.

The piecewise-smooth (PS) models [14-15] aims to minimize the energy function Eq.(1) using the level set method. The contour $C(p)$: $R\rightarrow\Omega$ is implicitly represented by a level set function $\phi(x)$: $\Omega\rightarrow R$, i.e., $C = \{x\in\Omega \mid \phi(x) = 0\}$. By approximating the image with two smooth functions $u^+(x)$ and $u^-(x)$ in the sub-regions $\Omega^+ = \{x\in\Omega: \phi(x) > 0\}$ and $\Omega^- = \{x\in\Omega: \phi(x) < 0\}$, respectively, the energy functional of the PS model is defined as follows

$$E^{PS}(u^+,u^-,\phi) = \int_\Omega (u^+ - I)^2 H(\phi)dx + \mu\int_\Omega |\nabla u^+|^2 H(\phi)dx + \int_\Omega (u^- - I)^2 (1-H(\phi))dx \\ + \mu\int_\Omega |\nabla u^-|^2 (1-H(\phi))dx + v\int_\Omega |\nabla H(\phi)|dx \tag{3}$$

where $H(\bullet)$ is a Heaviside functional defined as $H(z) = \begin{cases} 1, z \geq 0 \\ 0, z < 0 \end{cases}$, $u^+$ and $u^-$ must be obtained by solving the two damped Poisson equations before each iteration of the level set function, and the computational cost is very high. Moreover, $u^+$ and $u^-$ must be extended to the whole image domain, which is very difficult to implement in practice. More details about the implementation of the PS model can be found in [14].

The CV and PS energy functionals are non-convex with respect to the level set function $\phi$, and therefore they may fall into the local minima [29]. The global CV and PS models have been proposed in [28] by adding a total variation term into the energy functional and by utilizing Chambolle's method [30] to yield the global minimizer. However, the global CV and PS models, like the original CV and PS models, still have some drawbacks, such as failures to segment an image with intensity inhomogeneity for the CV model and large computational cost for the PS model, because the two smooth functions $u^+(x)$ and $u^-(x)$ still need to be iteratively solved.

### III. LOCALLY STATISTICAL ACTIVE CONTOUR MODEL

#### A. *Statistical Model of Intensity Inhomogeneity*

Let $\Omega$ be the image domain, $b(x)$: $\Omega\rightarrow R$ be an unknown bias field, $I(x)$: $\Omega\rightarrow R$ be the given image, $J(x)$: $\Omega\rightarrow R$ be the true signal to be restored, and $n(x)$: $\Omega\rightarrow R$ be noise. We consider the following model of intensity

inhomogeneity [11][19]

$$I(x) = b(x)J(x) + n(x) \qquad (4)$$

Suppose there are *N* objects in the image domain Ω, and denote by $\Omega_i$ the domain of the $i^{th}$ object. The true signal *J*(x) is often assumed to be piecewise constant within each object domain, i.e., $J(x) = c_i$ for $x \in \Omega_i$, where $c_i$ is a constant. The bias field *b* is often assumed to be smooth in the image domain Ω. The noise *n* is assumed to be Gaussian-distributed with zero mean and variance $\sigma_n^2$ [31]. Thus the image intensity can be approximated by a Gaussian distribution with mean $\mu_n$ and variance $\sigma_n^2$. However, using only one Gaussian model is not accurate enough to describe the statistical characteristics of image intensity. Often we use multiple Gaussian probability distributions to model the image intensity, with each distribution modeling the image intensity in each object domain. The distribution corresponding to the object domain $\Omega_i$ is [32]

$$p(I(y)|\theta_i) = \frac{1}{\sqrt{2\pi}\sigma_i} \exp\left(-\frac{(I(y)-\mu_i(x))^2}{2\sigma_i^2}\right) \qquad (5)$$

where $\mu_i(x)$ is the spatially varying mean, and $\sigma_i$ is the standard deviation. Since *b*(x) varies slowly, it can be assumed to be a constant in a small window [19]. Thus we can assume that $\mu_i(x) \approx b(x)c_i$. We symbolize $\theta_i = \{c_i, \sigma_i, b\}$ and $\boldsymbol{\theta} = \{\theta_i, i = 1, \ldots, N\}$ in our following discussions.

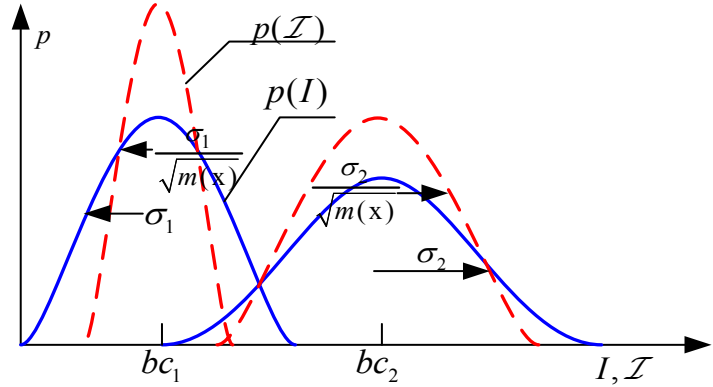

**Figure 1:** Distributions of adjacent regions in the original image intensity domain (blue solid curves) and in the transformed domain (red dashed curves).

As shown by the blue solid curves in Fig. 1, when *I*(x) falls into the tail of the distribution, misclassification will occur. One approach to reducing the overlapping tail is to compress the profile of the distribution while keeping its mean fixed [22] (see the red dashed curves in Fig. 1).

## B. Principle of the Proposed Method

For each position x in the image domain, we denote by $O_x$ a neighboring region center, i.e., $O_x = \{y \mid \|y - x\| \leq \rho\}$, where $\rho$ is the radius of the region $O_x$. Since there are $N$ non-overlapping objects in the image with $\Omega_i$ being the domain of the $i^{th}$ object, the whole image domain $\Omega$ can be represented as $\Omega = \cup_{i=1,\ldots,N}\Omega_i$ with $\Omega_i \cap \Omega_j = \varnothing, \forall\, i \neq j$. We define a mapping $T: I(x|\theta_i) \to \mathcal{I}(x|\theta_i)$ from the original image intensity domain $D(T)$ to another domain $R(T)$ as follows:

$$\mathcal{I}(x|\theta_i) = \frac{1}{m_i(x)} \sum_{y \in \Omega_i \cap O_x} I(y|\theta_i) \tag{6}$$

where $m_i(x) = \|\Omega_i \cap O_x\|$. The intensity of pixel x is assumed to be independently distributed [22]. Thus, $\forall \mathcal{I}(x|\theta_i) \in R(T)$, whereby the corresponding probability density function (PDF) is still a Gaussian [27], i.e., $\mathcal{I}(x|\theta_i) \sim N(\mu_i, \sigma_i^2 / m_i(x))$. Referring to the red dashed curves in Fig. 1, the overlapping tails of the distributions are suppressed to some extent.

Since the intensity inhomogeneity manifests itself as a smooth intensity variation across an image [19], we can assume that $I(y|\theta_i) \approx I(x|\theta_i)$, $\forall y \in \Omega_i \cap O_x$. Because the product of Gaussian PDFs is still Gaussian [27], we have

$$\prod_{y \in \Omega_i \cap O_x} p(I(y|\theta_i)) \approx p(I(x|\theta_i))^{m_i(x)} \propto N\left(\mu_i, \frac{\sigma_i^2}{m_i(x)}\right) \tag{7}$$

Therefore,

$$p(\mathcal{I}(x|\theta_i)) \approx \prod_{y \in \Omega_i \cap O_x} p(I(y|\theta_i)) \tag{8}$$

Let $D = \{\mathcal{I}(x|\theta_i), x \in \Omega, i = 1, \ldots, N\}$, we have the following likelihood function for the $i^{th}$ object [27]:

$$p(D|\theta_i) = \prod_{x \in \Omega} p(\mathcal{I}(x|\theta_i)) \tag{9}$$

We construct the following joint likelihood function

$$p(D|\boldsymbol{\theta}) = \prod_{i=1}^{N} p(D|\theta_i) = \prod_{i=1}^{N} \prod_{x \in \Omega} p(\mathcal{I}(x|\theta_i)) = \prod_{x \in \Omega} q(\mathcal{I}(x|\boldsymbol{\theta})) \tag{10}$$

where $\boldsymbol{\theta} = \{\theta_i, i = 1, \ldots, N\}$, and

$$q(\mathcal{I}(\mathrm{x}|\boldsymbol{\theta})) = \prod_{i=1}^{N} p(\mathcal{I}(\mathrm{x}|\theta_i)) \approx \prod_{i=1}^{N} \prod_{\mathrm{y} \in \Omega_i \cap O_\mathrm{x}} p(I(\mathrm{y}|\theta_i)) \tag{11}$$

Put Eq.(7) into Eq.(11), then Eq. (11) can be re-written as a univariate Gaussian distribution as follows:

$$q(\mathcal{I}(\mathrm{x}|\boldsymbol{\theta})) = \prod_{i=1}^{N} p(\mathcal{I}(\mathrm{x}|\theta_i)) \propto N(\mu, \gamma) \tag{12}$$

where

$$\mu = \gamma \sum_{i=1}^{N} \frac{m_i(\mathrm{x})\mu_i}{\sigma_i^2} \quad \text{and} \quad \gamma^{-1} = \sum_{i=1}^{N} \frac{m_i(\mathrm{x})}{\sigma_i^2} \tag{13}$$

Obviously, the joint likelihood function in Eq.(10) of each pixel (voxel) is composed of multiple classes of intensities; thus, by using Eq.(13), our model can yield a *soft classification*, satisfying the condition of the partial volume effect [33] (i.e., the intensity of each volume voxels is mixed from multiple classes [33]). Moreover, as can be seen from Eq.(6), the intensity in the transformed domain exploits the information about neighboring pixels belonging to the same class, so its classification result is less sensitive to noise and can result in a smoother object border.

We define the energy functional $l(\boldsymbol{\theta})$ as the log-likelihood function w.r.t. $p(D|\boldsymbol{\theta})$ in Eq.(10):

$$l(\boldsymbol{\theta}) \triangleq -\log p(D|\boldsymbol{\theta}) = \text{constant} - \sum_{i=1}^{N} \int_{\Omega} \int_{\Omega_i \cap O_\mathrm{x}} \log(p(I(\mathrm{y}|\theta_i))) d\mathrm{y} d\mathrm{x} \tag{14}$$

Let $\mathcal{K}_\rho(\mathrm{x},\mathrm{y})$ be the indicator function of region $O_\mathrm{x}$

$$\mathcal{K}_\rho(\mathrm{x},\mathrm{y}) = \begin{cases} 1, & \|\mathrm{y}-\mathrm{x}\| \leq \rho \\ 0, & \text{else.} \end{cases} \tag{15}$$

Using Eqs. (5) and (15), and eliminating the trivial constant term, $l(\boldsymbol{\theta})$ can be re-written as

$$l(\boldsymbol{\theta}) = \sum_{i=1}^{N} \int_{\Omega} \int_{\Omega_i} \mathcal{K}_\rho(\mathrm{x},\mathrm{y}) \left( \log(\sigma_i) + \frac{(I(\mathrm{y}) - b(\mathrm{x})c_i)^2}{2\sigma_i^2} \right) d\mathrm{y} d\mathrm{x} \tag{16}$$

## C. Energy Functional Formulation using the Level Set Method

One level set function $\phi$ can only represent two regions, inside and outside the contour $C$, as $\Omega^+ = inside(C) = \{\phi > 0\}$ and $\Omega^- = outside(C) = \{\phi < 0\}$, respectively. This is called the Two-Phase model. If there exist more than two different regions, two level set functions, $\phi_1$ and $\phi_2$, can be used to represent different regions based on the Four-Color Theorem [14] such that any two adjacent regions can be represented by different colors.

This is called the Four-Phase model. We define the phase indicators as follows:

$$\text{Two Phase:} \begin{cases} M_1(\Phi) = H(\phi) \\ M_2(\Phi) = 1 - H(\phi) \end{cases} \quad (17\text{-a})$$

$$\text{Four Phase:} \begin{cases} M_1(\Phi) = H(\phi_1)H(\phi_2) \\ M_2(\Phi) = H(\phi_1)(1 - H(\phi_2)) \\ M_3(\Phi) = (1 - H(\phi_1))H(\phi_2) \\ M_4(\Phi) = (1 - H(\phi_1))(1 - H(\phi_2)) \end{cases} \quad (17\text{-b})$$

where $H(\bullet)$ is the Heaviside functional, $\Phi$ represents the set of the level set functions such that $\Phi=\{\phi\}$ for the Two-Phase model, and $\Phi=\{\phi_1,\phi_2\}$ for the Four-Phase model. The energy functional $l(\theta)$ can then be re-written as

$$l(\boldsymbol{\theta},\Phi) = \sum_{i=1}^{N} \int_{\Omega} d_i(y) M_i(\Phi(y)) dy \quad (18)$$

where $d_i(y) \triangleq \int_{\Omega} \mathcal{K}_\rho(x,y)\left(\log(\sigma_i) + (I(y) - b(x)c_i)^2 / 2\sigma_i^2\right) dx$, $N = 2$ or 4.

## IV. RELATIONSHIP WITH OTHER MODELS

In the following, we explain the relationships of our model with five well-known active contour models, i.e., CV model [13], Geodesic active region (GAR) model [48][49], Local Region model (LRM) [41], LBF model [17], LIC model [19][47] and Local Gaussian distribution (LGD) model [46] in details.

We only consider the Two-Phase level set method in the following discussions. Similar discussions can be readily extended to the Four-Phase case. Let's revisit the data-fitting term of our proposed energy functional in Eq. (18). When the variance $\sigma_i^2 = 1, i = 1, 2$, and the bias field $b(x) = 1$, Eq. (18) can be rewritten as

$$l(\boldsymbol{\theta},\phi) = \frac{1}{2}\sum_{i=1}^{2} \iint \mathcal{K}_\rho(x,y)(I(y)-c_i)^2 M_i(\phi(y)) dx dy = \frac{1}{2}\sum_{i=1}^{2} \int (I(y)-c_i)^2 M_i(\phi(y)) dy \int \mathcal{K}_\rho(x,y) dx$$
$$= \frac{\pi\rho^2}{2}\sum_{i=1}^{2} \int (I(y)-c_i)^2 M_i(\phi(y)) dy = \frac{\pi\rho^2}{2}\left(\int (I(y)-c_1)^2 H(\phi(y)) dy + \int (I(y)-c_2)^2 (1-H(\phi(y))) dy\right) \quad (19)$$

This data-fitting term is similar to that of the well-known CV model [13] except for the trivial constant $\pi\rho^2/2$. Thus, our energy functional is a generalized version of the CV model, which uses two constants $c_1$ and $c_2$ which are the average intensities in different regions to approximate the image intensity in the regions

$\Omega^+=\{\phi(x)>0\}$ and $\Omega^-=\{\phi(x)<0\}$, respectively.

However, the average intensity in a large region cannot fit the intensity well when the intensity of the image is inhomogeneous. The boundary information can handle inhomogeneous intensity well. The GAR model [48][49] combines the region and boundary information into an energy functional. Different from the CV model only using the mean of region in Eq.(19), the region part in GAR model also considers the variance of region.

LRM [41] can handle the inhomogeneous intensity by using spatially varying means and variances to replace the constant mean and variance. However, LRM directly introduces a Gaussian kernel to compute the varying means and variances which is inconsistent with theory as pointed by [46].

The data-fitting term of the energy functional in the LBF model is as follows:

$$E_{data}^{LBF}(\phi, f_1, f_2) = \iint K_\sigma(y-x)|I(x)-f_1(y)|^2 H(\phi(x))dxdy + \iint K_\sigma(y-x)|I(x)-f_2(y)|^2 (1-H(\phi(x)))dxdy \quad (20)$$

where $K_\sigma(\bullet)$ is a truncated Gaussian kernel with standard deviation $\sigma$, which satisfies $\int K_\sigma(x)dx=1$, and $f_1$ and $f_2$ are two smooth functions to fit the image intensity in the regions $\Omega^+=\{\phi(x) > 0\}$ and $\Omega^-=\{\phi(x) < 0\}$, respectively. Minimizing $E_{data}^{LBF}(\phi, f_1, f_2)$, we can obtain that $f_1$ and $f_2$ are the weighted average image intensities in a Gaussian window inside and outside the contour, respectively. This is why the LBF model can handle images with intensity inhomogeneity well.

It is worth noting that our proposed model can also be seen as a generalization of the LBF model [17][18] and LIC model [19][47]. However, their intrinsic principles are different. If we set $\sigma_i^2 = 1$, $bc_i = f_i$, $i=1, 2$, in Eq. (18), and use a truncated Gaussian kernel $\mathcal{K}_\rho$ with standard deviation $\rho$ satisfying $\int \mathcal{K}_\rho(x)dx=1$ to replace the original constant kernel, then Eq. (18) can also be written as

$$\begin{aligned}l(\boldsymbol{\theta},\phi) &= \frac{1}{2}\sum_{i=1}^{2}\iint \mathcal{K}_\rho(x,y)(I(y)-f_i(x))^2 M_i(\phi(y))dxdy \\ &= \frac{1}{2}\sum_{i=1}^{2}\iint \mathcal{K}_\rho(x,y)(I(x)-f_i(y))^2 M_i(\phi(x))dxdy \\ &= \frac{1}{2}\sum_{i=1}^{2}\iint \mathcal{K}_\rho(y-x)(I(x)-f_i(y))^2 M_i(\phi(x))dxdy \\ &= \frac{1}{2}\left(\begin{aligned}&\iint \mathcal{K}_\rho(y-x)(I(x)-f_1(y))^2 H(\phi(x))dxdy \\ &+\iint \mathcal{K}_\rho(y-x)(I(x)-f_2(y))^2 (1-H(\phi(x)))dxdy\end{aligned}\right)\end{aligned} \quad (21)$$

This is the same as $E_{data}^{LBF}(\phi, f_1, f_2)$, except for the trivial constant 1/2. If we set $f_i = bc_i$, $i = 1, 2$ in Eq. (21), it is the energy functional of the LIC model [19][47]. It should be noted that in the theoretical analysis of our method in Section III.B, the constant kernel is reasonably used as a local region indicator. However, the LBF model and LIC model use Gaussian kernel as the locally spatially weighted function to relate the pixel x and its neighboring pixel y. The closer y is to x, the larger the weight is assigned, representing the higher similarity between the intensities of pixels y and x. Therefore, the principles among our method, the LBF model [17] and the LIC model [19] are different.

The energy functional of LGD model [46] is slightly different from the LBF model which replaces the L2-norm terms in Eq. (21) with $(I(x)-f_i(y))^2/\sigma_i^2(y), i=1,2$, where $f_i(y)$ and $\sigma_i(y)$ are spatially locally varying mean and variance of a Gaussian distribution. However, as pointed by [32], the spatially varying variance may be unstable due to the local property. Our model is different from LGD model due to the following reasons: First, the variances of the Gaussian distribution in our model is piecewise constant in each region, which is much more stable than the spatially varying variance in LGD model. Second, we use a constant kernel to indicator the local region which has a solid theoretical basis as explained in Section III.B. However, the Gaussian kernel is used in LGD model whose physical meaning is similar to the LBF model as we discussed above. Third, our model can be used for simultaneous segmentation and bias correction while the LGD model can be only used for segmentation.

## V. ENERGY MINIMIZATION AND LEVEL SET EVOLUTION FORMULATIONS

The minimization of $l(\theta,\Phi)$ with respect to (w.r.t.) each variable in $\theta = \{c_i, b, \sigma_i, i = 1,\ldots,N, N = 2, \text{ or } 4\}$ can be obtained by fixing other variables, yielding the closed forms of solutions, as described below.

### A. Closed Form Solutions for Different Variables

*Minimization w.r.t. $c_i$.* By fixing the other variables in Eq.(18), we obtain the minimizer of $c_i$, denoted by $\tilde{c}_i$, as follows:

$$\tilde{c}_i = \frac{\int (\mathcal{K}_\rho * b) I M_i(\Phi(y)) dy}{\int (\mathcal{K}_\rho * b^2) M_i(\Phi(y)) dy} \tag{22}$$

where $*$ denotes the convolution operator.

*Minimization w.r.t. b.* By fixing the other variables in Eq.(18), we obtain the minimizer of $b$, denoted by

$\tilde{b}$, as follows:

$$\tilde{b}(\mathrm{x}) = \frac{\sum_{i=1}^{N} \mathcal{K}_\rho * \left(IM_i\left(\Phi(\mathrm{x})\right)\right) \cdot \frac{c_i}{\sigma_i^2}}{\sum_{i=1}^{N} \mathcal{K}_\rho * M_i\left(\Phi(\mathrm{x})\right) \cdot \frac{c_i^2}{\sigma_i^2}} \qquad (23)$$

Note that $\tilde{b}$ is actually the normalized convolution [34], which naturally leads to a smooth approximation of the bias field $b$.

*Minimization w.r.t. $\sigma_i$.* By fixing the other variables in Eq.(18), we can obtain the minimizer of $\sigma_i$, denoted by $\tilde{\sigma}_i$, as follows

$$\tilde{\sigma}_i = \sqrt{\frac{\iint \mathcal{K}_\rho(\mathrm{y},\mathrm{x}) M_i\left(\Phi(\mathrm{y})\right) \left(I(\mathrm{y}) - b(\mathrm{x})c_i\right)^2 d\mathrm{y} d\mathrm{x}}{\iint \mathcal{K}_\rho(\mathrm{y},\mathrm{x}) M_i\left(\Phi(\mathrm{y})\right) d\mathrm{y} d\mathrm{x}}} \qquad (24)$$

For an explanation of how to derive the above solutions, please refer to **Appendices** A and B.

The obtained parameters $\tilde{\boldsymbol{\theta}} = \{\tilde{c}_i, \tilde{b}, \tilde{\sigma}_i, i=1,...,N, N=2, \text{or } 4\}$ are then put into Eq.(18), and the approximated $d_i$ can be derived by $\tilde{d}_i \triangleq \int_\Omega \mathcal{K}_\rho(\mathrm{x},\mathrm{y})\left(\log(\tilde{\sigma}_i) + \left(I(\mathrm{y}) - \tilde{b}(\mathrm{x})\tilde{c}_i\right)^2 / 2\tilde{\sigma}_i^2\right) d\mathrm{x}$.

## B. Two-Phase Level Set Evolution Formulation

Minimizing the energy functional $l(\tilde{\boldsymbol{\theta}}, \Phi)$ w.r.t. $\phi$, we have the corresponding gradient descent formulation as follows:

$$\frac{\partial \phi}{\partial t} = -\frac{\partial E(\tilde{\boldsymbol{\theta}}, \Phi)}{\partial \phi} = (\tilde{d}_2 - \tilde{d}_1)\delta(\phi) \qquad (25)$$

where $\delta(\phi)$ is the Dirac functional.

In order to keep the numerical implementation stable, the level set function should be regularized during the iteration of Eq.(25). Li *et al*. [35] proposed a signed distance-regularization formulation to regularize the level set function. However, as indicated by Xie [37], Li *et al*.'s method can produce some unnecessary valleys and peaks, which makes the level evolution easy to fall into some local minima. Li *et al*. [35] also proposed another improved signed distance-regularization formulation, but we found that the level set evolution is unstable if we use it in our formulation. The reason may be that we use different approximations for the Heaviside functional and the Dirac functional. In this paper we propose a new and simple method to regularize the level set function during iteration, which can make the evolution stable. After each iteration of level set evolution, we diffuse the level set function using the following formulation:

$$\phi^{n+1} = \phi^n + \Delta t_2 \cdot \nabla^2 \phi^n \tag{26}$$

where $\phi^n$ represents the level set function yielded by Eq. (25) during the $n^{th}$ iteration, $\nabla^2$ represents the Laplacian operator, and $\Delta t_2$ represents the diffusion strength. The $\phi^{n+1}$ in Eq. (26) can also be approximated by $\phi^{n+1}=K*\phi^n$, where K is either a Gaussian kernel [24] or a constant kernel [23].

## C. Four-Phase Level Set Evolution Formulation

Minimizing the energy functional $l(\tilde{\boldsymbol{\theta}},\Phi)$ with respect to $\phi_1$ and $\phi_2$, respectively, we can have the corresponding gradient-descent formulations as follows:

$$\begin{cases} \dfrac{\partial \phi_1}{\partial t} = -\left[(\tilde{d}_1 - \tilde{d}_2 - \tilde{d}_3 + \tilde{d}_4)H(\phi_2) + \tilde{d}_2 - \tilde{d}_4\right]\delta(\phi_1) \\ \dfrac{\partial \phi_2}{\partial t} = -\left[(\tilde{d}_1 - \tilde{d}_2 - \tilde{d}_3 + \tilde{d}_4)H(\phi_1) + \tilde{d}_3 - \tilde{d}_4\right]\delta(\phi_2) \end{cases} \tag{27}$$

Similar to the Two-Phase case, the level set functions are regularized by the following formula after each iteration:

$$\phi_i^{n+1} = \phi_i^n + \Delta t_2 \cdot \nabla^2 \phi_i^n, i=1,2 \tag{28}$$

## D. Numerical Implementation

In the level set evolution of Eqs. (25) and (27), we only need to approximate the temporal derivative as a forward difference because there are no partial derivatives. The Laplacian operator $\nabla^2 \phi$ is approximated by $\nabla^2 \phi \approx K*\phi$, where $*$ is a convolution operator, and K is a kernel defined as $K=\begin{bmatrix} 0 & 1 & 0 \\ 1 & -4 & 1 \\ 0 & 1 & 0 \end{bmatrix}$ [37]. Therefore, the solution of the diffusion equation Eq.(26) can be discretized as follows:

$$\phi^{n+1} = \phi^n * \begin{bmatrix} 0 & \Delta t_2 & 0 \\ \Delta t_2 & 1-4\Delta t_2 & \Delta t_2 \\ 0 & \Delta t_2 & 0 \end{bmatrix} \tag{29}$$

The standard Von Neumann analysis [43] can be used to analyze the stability for the time step $\Delta t_2$. Putting

$\phi^n(i,j)=r^n e^{I(i\xi_1+j\xi_2)}$ into Eq. (29), where $I=\sqrt{-1}$ denotes the imaginary unit, we obtain the amplification factor as follows:

$$r = 1 + 2\Delta t_2 \cdot \left[\cos(\xi_1) + \cos(\xi_2) - 2\right] \tag{30}$$

Therefore, we have $1-8\Delta t_2 \leq r \leq 1$. By solving the inequality $|1-8\Delta t_2|\leq 1$, we have:

$$0 \leq \Delta t_2 \leq 0.25 \tag{31}$$

The Heaviside functional $H(z)$ is approximated by a smooth function $H_\varepsilon(z)$ as

$$H_\varepsilon(z) = \frac{1}{2}\left[1 + \frac{2}{\pi}\arctan\left(\frac{z}{\varepsilon}\right)\right], z \in R \tag{32}$$

where $\varepsilon > 0$ is a constant. The Dirac functional $\delta(z)$ is approximated by $\delta_\varepsilon(z)$ as follows

$$\delta_\varepsilon(z) = \frac{d(H_\varepsilon(z))}{dz} = \frac{1}{\pi}\frac{\varepsilon}{\varepsilon^2 + z^2}, z \in R \tag{33}$$

The profile of $\delta_\varepsilon(z)$ has a large support, which acts on all level curves of the level set function. This can allow the level set evolution to reach a global minimum easily [13].

Finally, based on the above description of our algorithm, the procedures for our proposed algorithm are summarized in Table I.

**TABLE I:** PROCEDURES OF OUR ALGORITHM

| **Algorithm of Locally Statistical ACM** |
|---|
| **1.** Initialization: $\tilde{b}=1$, $\tilde{\sigma}_i = i$, $i = 1,\ldots,N$, and the level set function $\phi_i^n = \phi_i^0$, $i=1,\ldots,1$ or 2, $n = 1$; |
| **2.** Update $c_i$ to $\tilde{c}_i$, $i = 1,\ldots,N$, by Eq.(22); |
| **3.** Update $b$ to $\tilde{b}$ by Eq. (23); |
| **4.** Update $\sigma_i$ to $\tilde{\sigma}_i$, $i = 1,\ldots,N$, by Eq.(24); |
| **5.** Update $d_i$, $i=1,\ldots,N$ to $\tilde{d}_i$, $i = 1,\ldots,N$, respectively; |
| **6.** Evolve the level set function according to Eq.(25) or (26) once; |
| **7.** Regularize the level set function according to Eq.(29). |
| **8.** If $\phi_i^{n+1}$ satisfies the stationary condition, stop; otherwise, $n = n+1$ and return to Step **2**. |

## VI. EXPERIMENTAL RESULTS

In this section, we compare our method with the CV model [13], the global CV (GCV) model [28] (the code was downloaded from [39]), the LRB model [21] (the code was downloaded from [40]), the LIC model [19], and the LBF model [17][18] (the code was downloaded from [38]), which are representative and state-of-the-art ACMs for image segmentation. The Matlab source codes and some examples of the proposed method can be downloaded at http://www.comp.polyu.edu.hk/~cslzhang/LSACM/LSACM.htm.

It is easy to choose the parameters in our model. We initialize $\tilde{b}=1$, $\tilde{\sigma}_i = i$, $i = 1, \ldots, N$, and then the initialization of $\tilde{c}_i$, $i = 1, \ldots, N$, can be calculated by Eq.(22). We set the time step for level set evolution as $\Delta t = 1$, the time step for the regularization as $\Delta t_2 = 0.1$, and $\varepsilon = 1$ for all the experiments except for Figs. 7, 8, and 9, for which we set $\Delta t_2 = 0.01$. Our method is stable for a wide range of $\rho$, e.g., $5 < \rho < 25$. In most cases, we set $\rho = 6$. A small $\rho$ will make the computation in each iteration more efficient, but the convergence is slower. On the other hand, a large $\rho$ will increase the computational burden in each iteration, while the convergence rate can be increased because information from larger regions is used. Therefore, the total computation burden is comparable for different $\rho$.

In Section V-A, we will compare our method with the CV model and the GCV model for segmenting two synthetic images and two real vessel images with intensity inhomogeneity. In Section V-B, we compare our method with the LBF model, the LIC model and the LRB model for segmenting a synthetic image and two real MR brain images with severe intensity inhomogeneity. Sections V-A and V-B focus on the Two-Phase level set model, and in Section V-C we test our method for simultaneous segmentation and bias correction. We first use a synthetic image with the ground-truth bias field to qualitatively evaluate our method in comparison with the LIC model. Then, we compare our method with the LIC model on a real 3T MR brain image with different initial contours, and apply our method to two 7T MR brain images. At last, in Section V-D, we quantitatively compare our method with the CV model, the GCV model, the LRB model, the LBF model, and the LIC model by segmenting five synthetic images with gradually increasing strength of intensity inhomogeneity.

### A. *Comparisons with the CV Model [13] and the GCV Model [28]*

Both the CV model [13] and the GCV model [28] assume that the image intensity is piecewise constant and use the global intensity means to fit the image intensity. Therefore, they do not perform well in images with intensity inhomogeneity. In this section, we compare our method with the CV model and the GCV model by applying them to some real and synthetic images with intensity inhomogeneity in order to demonstrate the superior performance of our method to them for images with intensity inhomogeneity.

Fig. 2 demonstrates the segmentation results on two synthetic images and two real vessel images with intensity inhomogeneity by using the CV model [13], the GCV model [28], and our method, respectively. It can be seen that our method yields satisfactory segmentation results because we consider and exploit the image local region information, which can better separate the object from background. For the CV model and the GCV model, they use the global region information for segmentation and thus result in severe misclassification on these images because there exist severe overlaps of intensity between the objects and background. Using the global mean information cannot discriminate the overlapping intensity.

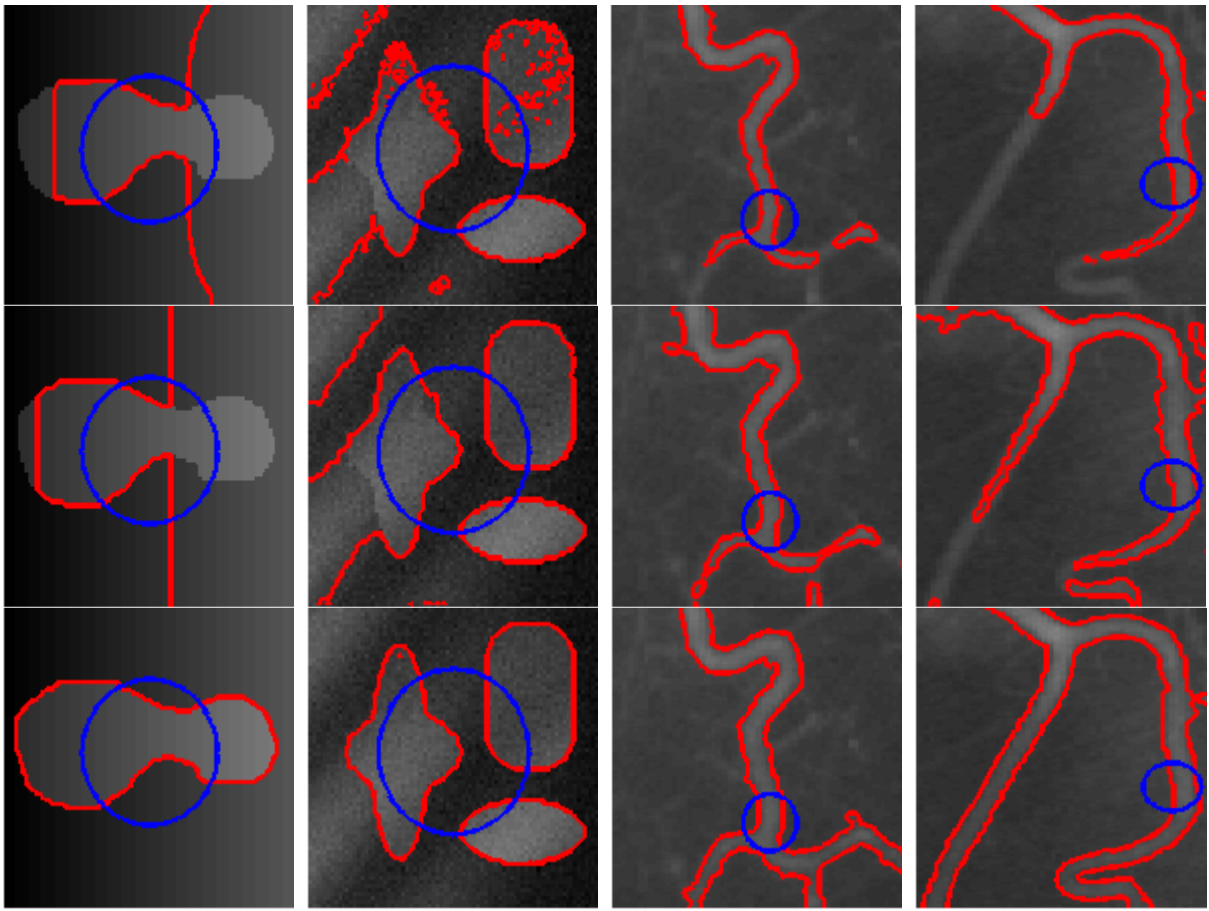

**Figure 2:** Segmentation results on two synthetic images and two real vessel images with intensity inhomogeneity (downloaded from [38]). From top row to bottom row: segmentation results by the CV model [13], the GCV model [28] and our method. The blue circle represents the initial contour and the red lines represent the final segmentation contour. We set $\rho = 12$ for the left image and $\rho = 6$ for the other images.

*B. Comparisons with the LBF Model [17][18], LIC Model [19] and the LRB Model [21]*

In this section, we use one synthetic image and two real MR images with severe intensity inhomogeneity to demonstrate the superior performance of our method to the LBF model, the LIC model and the LRB model, which all use local means to fit the image intensity and perform well in images with light intensity inhomogeneity. The LBF model [17][18] uses the local intensity mean to fit the measured image, and thus it can yield better segmentation results than CV and GCV models on images with intensity inhomogeneity. However, if the intensity inhomogeneity is severe, using only the local mean information may fail to discriminate the intensity between an object and its background, leading to inaccurate segmentation. Some experimental results are shown in the second column of Fig. 3. Similar drawbacks exist for the LIC model [19] (refer to the third column of Fig. 3) and LRB model [21]. Moreover, the localized version of the Heaviside functional and the Dirac functional used by the LRB model makes the level set evolution easily fall into local minima [13], as can be seen in the fourth column of Fig. 3. Our method can produce much better segmentation results because it considers the statistical information in a transformed domain, where the intensity of the object and background is less overlapping than in the original domain, making our method have a very strong discriminative capability for the object and background (refer to the left column of Fig. 3).

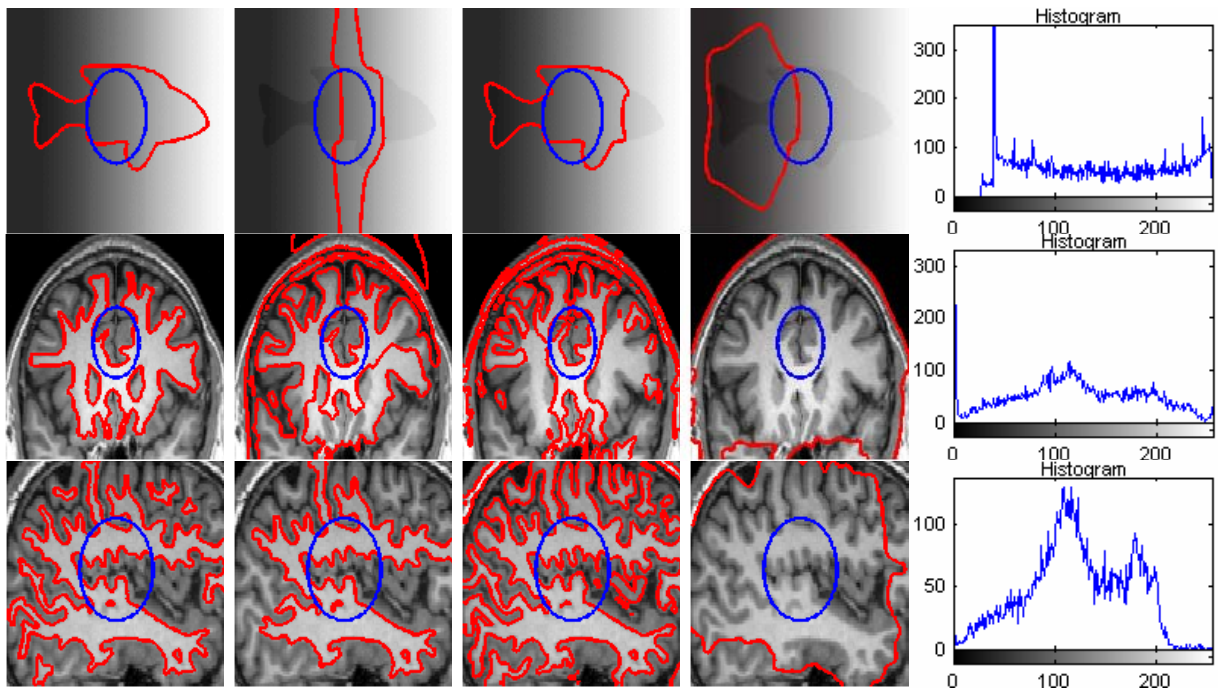

**Figure 3:** Segmentation on a synthetic image (1st row) and two real MR images (2nd and 3rd rows) with severe intensity inhomogeneity. From left to right: segmentation results using our method, the LBF model [17][18], the LIC model [19] and the LRB model [21], and the histograms of image intensity. The blue circles represent the initial contours, and the right lines represent the final segmentation. We set $\rho = 6$ for the first two images and $\rho = 20$ for the last image.

## C. Application to Simultaneous Segmentation and Bias Correction

In this section, we apply our ACM for simultaneous segmentation and bias correction, especially for MR images. We compare our method with the state-of-the-art LIC model [19] because only the LIC model is applicable for simultaneous segmentation and bias correction while the other models (i.e., the CV model, the GCV model, the LBF model and the LRB model) can only be used for segmentation.

Fig. 4 shows the results by the two competing methods on synthetic images corrupted with additive Gaussian noises of different levels. The evaluation criteria are the final segmentation results and the similarity between the estimated bias field and the ground-truth bias. The top-left image in Fig. 4 was added with Gaussian white noise of zero mean and unit standard deviation. Since the noise level is low, both the two models can yield satisfactory segmentation results, while our method outperforms a little the LIC model. However, with the increase in noise level, the segmentation results by the LIC method will become very noisy, as can be seen in the second image, second row, Fig. 4, where the standard deviation of added Gaussian noise is 5. Meanwhile, the segmentation result using our method is much better. This is because our model considers the different probability distributions of various objects. The estimated bias fields by the two methods are normalized and shown in the right two columns of Fig. 4. One can see that the estimated bias field by our method is visually much more similar to the ground-truth bias than the LIC method.

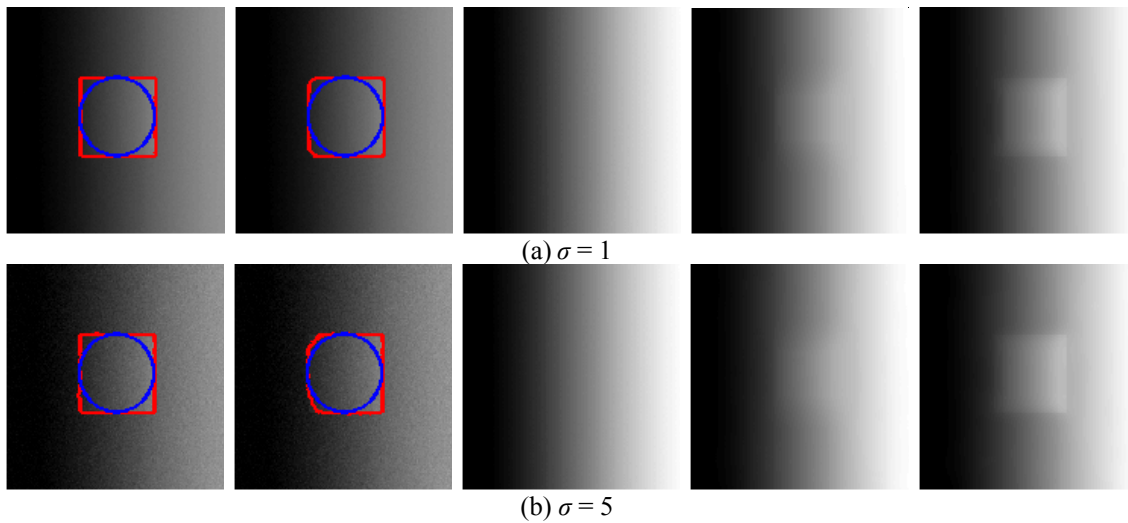

(a) $\sigma = 1$

(b) $\sigma = 5$

**Figure 4:** Experiments on a synthetic image with different additive Gaussian noises. The test images in the two rows are corrupted with Gaussian white noises of standard deviations $\sigma = 1$ and $\sigma = 5$, respectively. From left to right: the segmentation results by our method, the segmentation results by the LIC model [19], the ground-truth bias field, the estimated bias field using our method, and the estimated bias field using the LIC model [19]. We set $\rho = 10$ for the experiments.

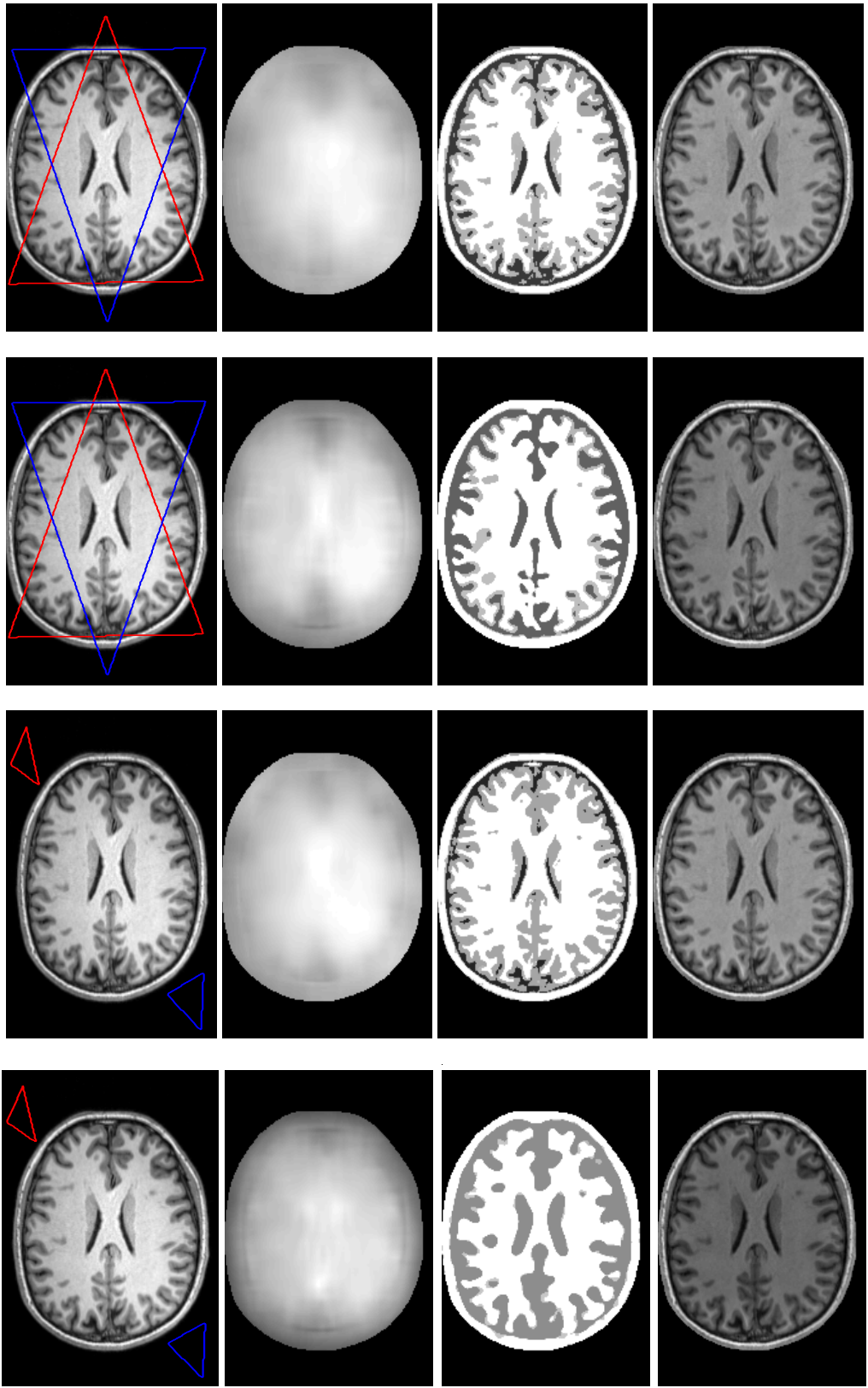

**Figure 5:** From left to right: the initializations of level set functions $\phi_1$ and $\phi_2$ (the red line represents the initial zero level set of $\phi_1$, while the blue line represents the initial zero level set of $\phi_2$), estimated bias fields, tissue-classification results, and bias-corrected images. The 1$^{st}$ and 3$^{rd}$ rows are the results by our method with two different initializations, while the 2$^{nd}$ and 4$^{th}$ rows are the results by the LIC model [19] with the same two initializations. We set $\rho = 10$ for all the experiments.

Fig. 5 shows the joint segmentation and bias-correction results on a 3T MRI image, which has four classes of tissues: whiter matter (WM), gray matter (GM), cerebrospinal fluid (CSF), and the background. Because one level set function can only represent two classes of tissues, we need to evolve two level set functions $\phi_1$ and $\phi_2$ according to Eq.(24) for four classes of tissues [14]. It can be seen that the tissue segmentation results by our method (the 1st and 3rd rows in Fig. 5) are much more accurate than those by the LIC method (the 2nd and 4th rows in Fig. 5). It should be noted that it is very easy to initialize the level set functions in our method. The initial contours can be set inside, outside or across the object boundary. The two initializations in the 1st and 3rd rows are very different, but the final segmentation results by our method are very similar. This demonstrates the robustness of our method to initialization.

Finally, we show in Fig. 6 the bias-correction results using our method on two 7T MRI images. This experiment aims to obtain the bias corrected MRI images. The original images, estimated bias fields, and the bias-corrected images are shown in the left, middle, and right columns, respectively. It can be clearly seen that the image quality is significantly improved by our method. Some regions (inside the red circles) whose intensity contrast is too low to be identified are able to be distinguished clearly after the bias correction.

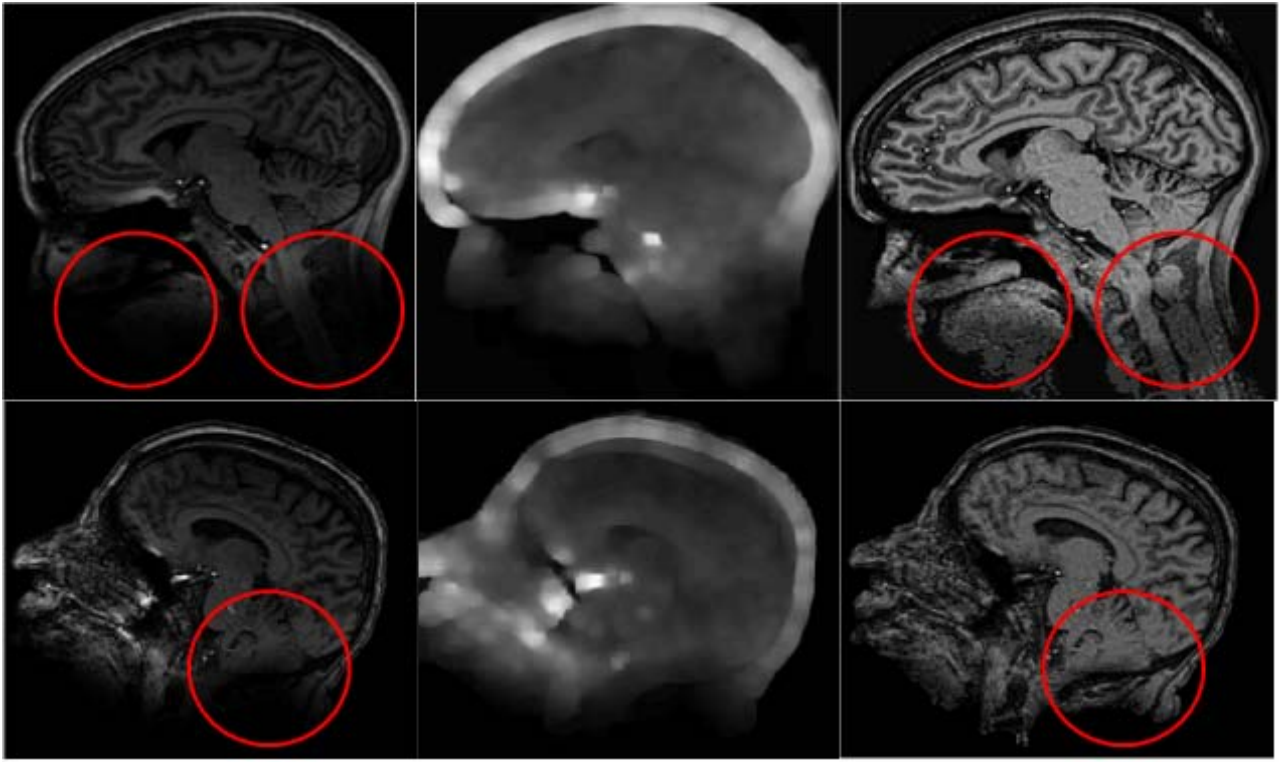

**Figure 6:** Bias-correction results on two 7T MRI images using the proposed method. From left to right: original images, estimated bias fields, and the bias-corrected images. We set $\rho = 10$ for both experiments.

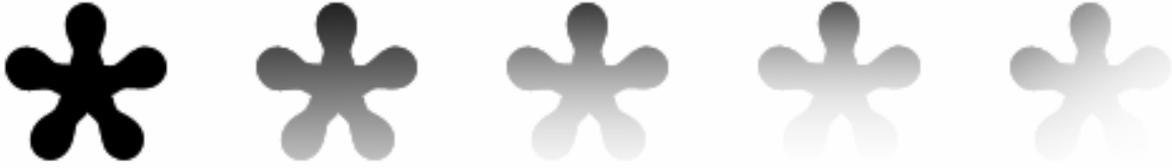

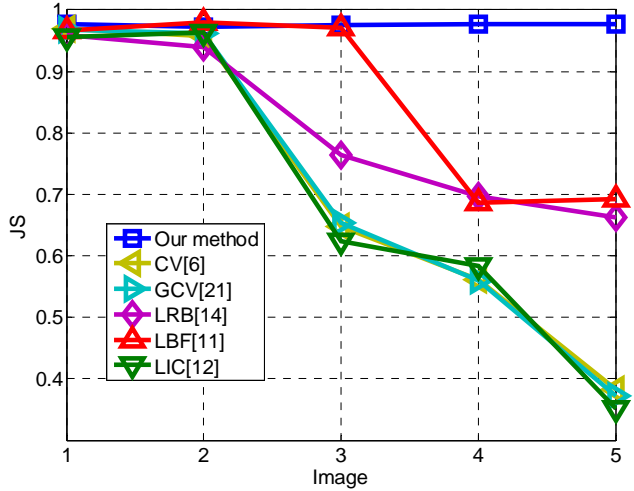

**Figure 7:** Quantitative comparisons among our method, the CV model [13], the GCV model [28], the LRB model [21], the LBF model [17][18], and the LIC model [19]. Top row: tested images, where the strength of intensity inhomegeneity is gradually increased from left to right. Bottom row: the corresponding *JS* values yielded by the competing methods on the five images. We set $\rho$ =10.5 for all the experiments.

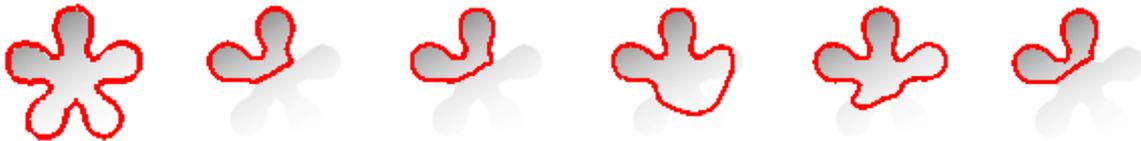

**Figure 8.** From left to right: segmentation results on image with strong intensity inhomegeneity by our method, the CV model [13], the GCV model [28], the LRB model [21], the LBF model[17][18], and the LIC model [19]. We set $\rho$ = 10.5 for all the experiments.

## D. *Quantitative Evaluation*

In this section, we use the Jaccard similarity (*JS*) [44] as an index to measure the segmentation accuracy and to quantitatively evaluate the segmentation performance of competing methods on images whose intensity inhomogeneity has different strength. The *JS* index between two regions $S_1$ and $S_2$ is calculated as $JS(S_1,S_2) = |S_1 \cap S_2|/|S_1 \cup S_2|$, which is the ratio between the intersectional area of $S_1$ and $S_2$ and their union. Obviously, the closer the *JS* value is to 1, the more similar $S_1$ is to $S_2$. In our experiments, $S_1$ is the segmented object region produced by the six competing methods (the CV model [13], the GCV model [28], the LRB model [21], the LBF model [17][18], the LIC model [19], and our proposed method), and $S_2$ is the ground-truth.

We test the six competing methods on five synthetic images with different intensity inhomogeneity. The *JS* values of the six methods are shown in the bottom image of Fig. 7. Obviously, the *JS* values obtained by our method have little difference for intensity inhomogeneity with different strength, which demonstrates that it is very robust to image intensity inhomogeneity. For the global region-based CV model and GCV model, when the strength of intensity inhomogeneity is not strong (refer to images 1 and 2 in Fig. 7), both of them can yield a high *JS* value. However, when the strength of inhomogeneity is not low (refer to images 3, 4, and 5), the performance of the CV model and the GCV model degenerates rapidly. For the local region-based models (i.e., the LRB model, the LBF model and the LIC model), when the strength of intensity inhomegeneity is strong (refer to images 4 and 5), the performance of these methods also degenerates severely. This is because using only the local region means cannot discriminate the object and the background satisfactorily when the intensity between the object and background overlaps severely. Our method gives consistently the highest *JS* values because it pursues the segmentation in a transformed domain where the intensity overlapping is much suppressed. The final segmentation results for the top right-most image in Fig. 7 by the six competing methods are shown in Fig. 8.

*E. Robustness to Initializations and Region Scale Parameters*

We use the tested image in Fig. 8 to demonstrate the robustness of our method to different level set initializations and region scale parameter $\rho$. We again use the *JS* index to measure segmentation accuracy. The top row in Fig. 9 demonstrates the segmentation results using different initial contours (the blue lines). We can see that there are no obvious visual differences for these segmentation results. (The *JS* values of these results correspond to the first six values shown in the bottom left figure of Fig. 9.) We apply 20 different initial contours for segmentation (the region scale parameter is set as $\rho = 10.5$) and compute the corresponding *JS* values. From the bottom-left figure of Fig. 9, we can see that the *JS* values change only from 0.97 to 0.98. These *JS* values clearly demonstrate that our method can yield very high and stable segmentation accuracies for different level set initializations. The bottom-right figure of Fig. 9 shows the *JS* values computed by changing the region scale parameter $\rho$ from 5.5 to 22.5. The high and stable *JS* values again demonstrate that our method is very robust to the region scale parameter is in a wide range.

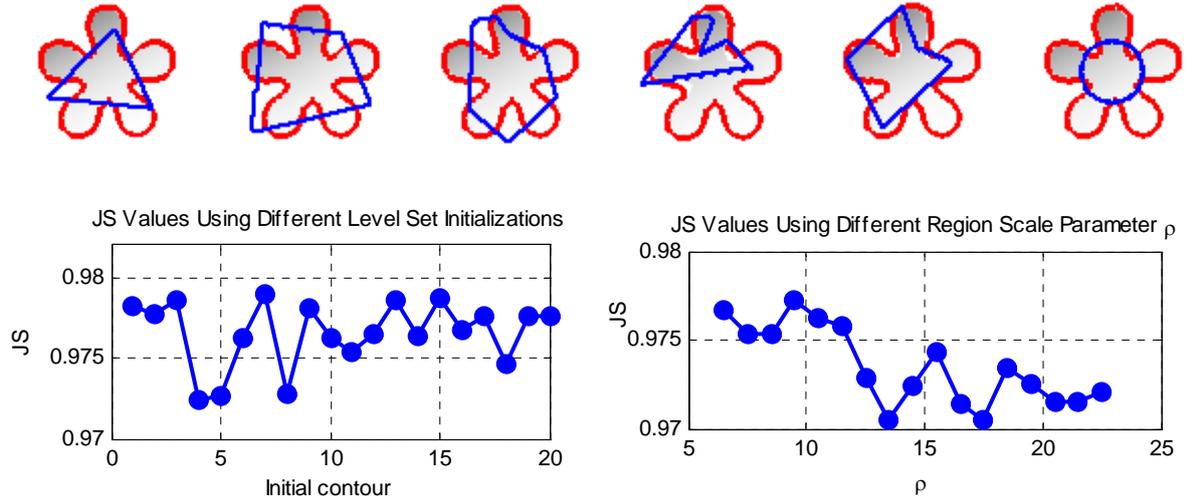

**Figure 9:** Segmentation accuracy by our method with different level set initializations and region scale parameter $\rho$. Top row: segmentation results with different level set initializations by setting $\rho = 10.5$. The blue lines represent different initial contours, and the red lines represent the final segmentation contour. Bottom row: the left figure shows the *JS* values for 20 different initial contours (the first six values are for the six segmentations in the top row of Fig. 9), while the right figure shows the *JS* values for different region scale parameter $\rho$. The initial contour is the same as that in the top right-most image.

## VII. CONCLUSION

In this paper, we presented a locally statistical active contour model for segmenting images with intensity inhomogeneity. Our method combines the information about neighboring pixels belonging to the same class, which makes it strong to separate the desired objects from background. Moreover, the proposed method yields a soft segmentation, which can, to some extent, satisfy the condition of the partial volume effect. In addition, the segmentation results are very robust to the initialization of the level set function, making it useful for automatic applications. Comparisons with five representative and state-of-the-art methods on synthetic and real images have demonstrated the effectiveness and the advantages of the proposed algorithm.

## APPENDIX A

In Eq. (18), we assume that the optimal $c_i$ is $\tilde{c}_i$ and we add variation $\eta_i$ to the variable $\tilde{c}_i$ such that $c_i = \tilde{c}_i + \varepsilon \eta_i$. Keeping other variables except for $c_i$ fixed, differentiating with respect to $c_i$ and letting $\varepsilon \to 0^+$, we have

$$\frac{\delta E}{\delta \tilde{c}_i} = \lim_{\varepsilon \to 0^+} \frac{dE(c_i)}{d\varepsilon}$$
$$= -\eta_i \iint \mathcal{K}_\rho(\mathrm{x},\mathrm{y})\left(I(\mathrm{y})b(\mathrm{x}) - b^2(\mathrm{x})\tilde{c}_i\right)/\sigma_i^2 M_i(\Phi(\mathrm{y}))\,d\mathrm{x}d\mathrm{y} = 0$$

Therefore, we have

$$\iint \mathcal{K}_\rho(\mathrm{x},\mathrm{y})\left(I(\mathrm{y})b(\mathrm{x}) - b^2(\mathrm{x})\tilde{c}_i\right)/\sigma_i^2 M_i(\Phi(\mathrm{y}))\,d\mathrm{x}d\mathrm{y} = 0$$

From the above equation, we obtain

$$\tilde{c}_i = \frac{\int (\mathcal{K}_\rho * b) I M_i(\Phi(\mathrm{y}))\,d\mathrm{y}}{\int (\mathcal{K}_\rho * b^2) M_i(\Phi(\mathrm{y}))\,d\mathrm{y}}$$

The $\tilde{\sigma}_i$ in Eq. (23) can be obtained using a similar method.

# APPENDIX B

In Eq. (18), we assume that the optimal smoothing function $b(x)$ is $\tilde{b}(x)$. Then, we add a variation function $\eta(x)$ to the variable $\tilde{b}(x)$ such that $b(x) = \tilde{b}(x) + \varepsilon\eta(x)$. Differentiating with respect to $b$ and letting $\varepsilon \to 0^+$, we have

$$\frac{\delta E}{\delta \tilde{b}} = \lim_{\varepsilon \to 0^+} \frac{dE(b)}{d\varepsilon}$$

$$= \int \sum_{i=1}^{N} \int \mathcal{K}_\rho(x,y) M_i(\Phi(y)) \left( I(y) - \tilde{b}(x)c_i \right) / \sigma_i^2 c_i dy \, \eta(x) dx$$

Therefore, we have the Euler-Lagrange equation

$$\sum_{i=1}^{N} \int \mathcal{K}_\rho(x,y) M_i(\Phi(y)) \left( I(y) - \tilde{b}(x)c_i \right) / \sigma_i^2 c_i dy = 0$$

Finally, we obtain

$$\tilde{b}(x) = \frac{\sum_{i=1}^{N} \mathcal{K}_\rho * \left( I M_i(\Phi(x)) \right) \cdot \frac{c_i}{\sigma_i^2}}{\sum_{i=1}^{N} \mathcal{K}_\rho * M_i(\Phi(x)) \cdot \frac{c_i^2}{\sigma_i^2}}$$